\documentclass[letterpaper, 10 pt, conference]{ieeeconf}  

\IEEEoverridecommandlockouts                              

\usepackage{amsmath}
\usepackage{amssymb}
\usepackage{mathptmx}
\usepackage{url}
\usepackage{times}

\usepackage{xcolor}

\usepackage{multicol}
\usepackage[bookmarks=true]{hyperref}
\usepackage[ruled, vlined, linesnumbered]{algorithm2e}
\usepackage{subcaption}
\usepackage{graphicx}

\title{\LARGE \bf
Conditional Task and Motion Planning through an Effort-Based Approach
}

\author{Nicola Castaman$^{1}$, Elisa Tosello$^{2}$, and Enrico Pagello$^{3}$
\thanks{$^{1}$Nicola Castaman is with IAS-Lab, Dept. of Information Engineering,
        University of Padova, Padova, Italy and IT+Robotics srl, Vicenza, Italy
        {\tt\small castaman@dei.unipd.it}}%
\thanks{$^{2}$Elisa Tosello is with IAS-Lab, Dept. of Information Engineering,
        University of Padova, Padova, Italy
        {\tt\small toselloe@dei.unipd.it}}%
\thanks{$^{3}$Enrico Pagello is with IAS-Lab, Dept. of Information Engineering,
        University of Padova, Padova, Italy and IT+Robotics srl, Vicenza, Italy
        {\tt\small epv@dei.unipd.it}}%
}

\begin{document}

\maketitle
\thispagestyle{empty}
\pagestyle{empty}

\begin{abstract}

This paper proposes a preliminary work on a Conditional Task and Motion Planning algorithm able to find a plan that minimizes robot efforts while solving assigned tasks. Unlike most of the existing approaches that replan a path only when it becomes unfeasible (e.g., no collision-free paths exist), the proposed algorithm takes into consideration a replanning procedure whenever an effort-saving is possible. The effort is here considered as the execution time, but it is extensible to the robot energy consumption. The computed plan is both conditional and dynamically adaptable to the unexpected environmental changes. Based on the theoretical analysis of the algorithm, authors expect their proposal to be complete and scalable. In progress experiments aim to prove this investigation.

\end{abstract}

\section{Introduction}

\begin{figure}[t!]
        \centering\includegraphics[scale=0.75]{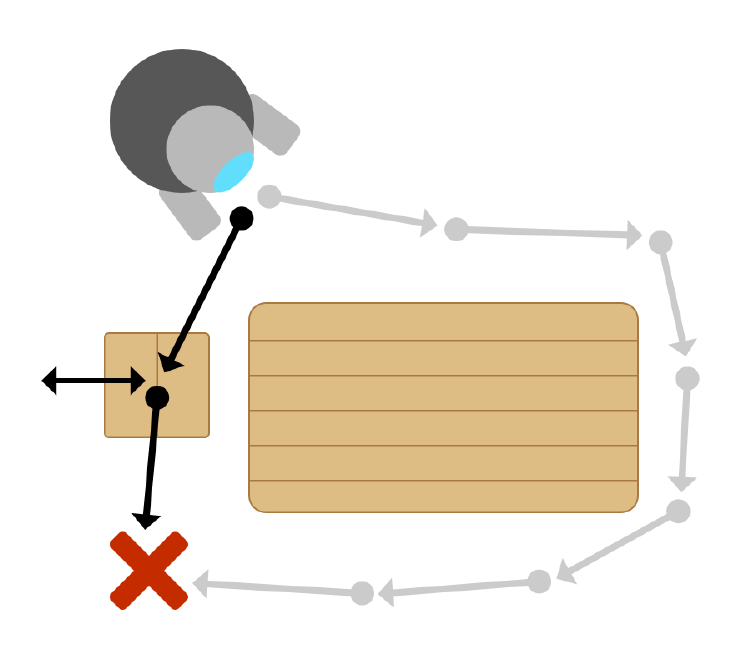}
        \caption{A Navigation Among Movable Obstacles problem: while navigating towards a goal, the robot takes into consideration the possibility of moving movable obstacles.}
        \label{fig:navigation}
\end{figure}

Let a human assign a task to a robot. e.g., the pick of a can of coke from a cluttered table or the resolution of a Navigation Among Movable Obstacles (NAMO)~\cite{dogar2011framework, levihn2013hierarchical} problem (see Figure~\ref{fig:navigation}). In order to achieve these targets, the robot needs to fulfill high-level task planning in conjunction with low-level motion planning. As stated in~\cite{dantam2016incremental}, efficient algorithms exist to solve task and motion planning problems in isolation; however, their integration is still challenging in terms of generality, completeness, and scalability.

Authors provide a Task and Motion Planning (TAMP) system which combines a Fast-Forward (FF) task planner~\cite{hoffmann2001ff} and a revisited version of the Lazy Kinodynamic Motion Planning by Interior-Exterior Cell Exploration (L-KPIECE) motion planner~\cite{csucan2009kinodynamic}, as proposed in~\cite{castaman2016thesis,castaman2016sampling}. The goal is to give the possibility to any type of robot to use this system for solving any type of task, from manipulation to navigation, in an unknown, real-world scenario. The implementation aims to reflect the human behavior: humans take plans while efficiently managing their time and energy. 
In detail, given the actions that the robot can perform, together with its initial and final states, one feasible task plan, namely a \textit{reachability graph}, is computed by using FF (no optimality check is required). The corresponding motion plan is generated by using~\cite{castaman2016sampling}.
Based on the knowledge of the robot world at its current state, a collision checking is performed so that the final plan would take into consideration only those objects and sub-tasks that minimize robot efforts (the execution time in this case, but the energy used by the robot can also be considered). 
The robot starts moving. While acting, if failures or environment changes occur (e.g., objects are detected which obstruct the computed path), an online replanning routine is adopted. Starting from the robot current state, it regenerates the remaining task plan by evaluating and substituting those actions which preconditions are no longer feasible. Starting from the new task plan, the algorithm proposed in~\cite{castaman2016sampling} is invoked to find a new feasible set of motions.

\textit{Completeness.} Acting in the real world is non-deterministic: the planner has no complete and certain knowledge of the environment and actions can have unpredictable effects. This means that a robot should be able to dynamically adapt to changes and efficiently recover from failures while avoiding the explosion of the computed \textit{reachability graph} because of the new alternatives introduced. In the case of analysis, starting from a \textit{relax} task, i.e., a plan which ignores the delete lists of operators,
FF extracts an \textit{explicit solution} to the plan by using a search strategy called \textit{enforced hill-climbing} (EHC). This strategy does not use backtracking, meaning that some parts of the search space are lost, but if FF fails to find a solution using EHC by getting stuck in dead-ends, it switches to standard best-first searches (e.g., \textit{greedy} or \textit{A*} search) which expands \textit{all} search nodes by increasing order of goal distance evaluation. This recovery routine guarantees completeness. As stated in~\cite{csucan2009kinodynamic}, KPIECE is, instead, probabilistically complete. Moreover, the proposed planner handles known and unknown events through the online replanning routine until a maximum number of attempts is reached. This means that if the number of attempts tends to infinity, the probability of finding a plan, if one exists, will tend to one.

\textit{Scalability.}
The task planning problem is exponential in the number of sub-tasks and the motion planning problem is exponential in the number of collision objects populating the workspace. Indeed, multiple sub-tasks exist fulfilling the assigned task and different combinations of sub-tasks can bring to the same result. Moreover, when operating in the clutter, a robot has to decide which objects to move, where to place them, if moving them is necessary or convenient.  
Focusing on the task-level planning, FF employs an \textit{helpful actions pruning} that maintains in the \textit{relaxed} plan only those actions that are really useful in a state, so one can restrict the successors of any state to only those produced by members of the respective relaxed solution. 
Focusing on the motion-level planning, \cite{castaman2016sampling} is based on KPIECE because KPIECE uses the information collected during the planning process in order to decrease the amount of forward propagation it needs~\cite{csucan2009kinodynamic}. Moreover, by using the Lazy strategy, edges of the computed path are not immediately checked for collision: they are checked only when a better path to the goal is found. 
By eliminating the majority of collision checks and efficiently forward propagating the exploration during the planning, the proposed algorithm speeds up to convergence in complex problems, like that of TAMP, where collision checking is relatively expensive. It also requires restrained runtime and memory so that it can handle high dimensional systems with complex dynamics.

\section{Related Work}

Various researchers investigated the problem of combining task and motion planning. Most of them tried to combine the symbolic reasoning of task planning with the geometric reasoning of motion planning. 
Dornhege et al.~\cite{dornhege2012semantic,dornhege2013lazy}, for instance, proposed a system that calls the motion planner only to check the geometric feasibility of the planned tasks. Another example is the FFRob approach proposed by Garrett~\cite{garrett2015ffrob,garrett2016ffrob}: it integrates geometric information with the state-of-the-art Fast-Forward (FF)~\cite{hoffmann2001ff} task planner by sampling a fixed set of object poses and robot configurations and then planning with them by using FF. 
Benefits of FFRob are its probabilistic completeness and exponentially convergence. Its major limit is its focus: the approach solves only a particular class of pick-and-place problems. Moreover, it does not use planning to guide sampling.

Other approaches such as~\cite{srivastava2014combined,de2013towards,gharbi2015combining} still focus on manipulation tasks but they propose solutions based on Hierarchical Planning that evaluate task-level decisions by using low-level geometric-reasoning modules. In particular, Srivastava~\cite{srivastava2014combined} combines off-the-shelf task planners with an optimization-based motion planner~\cite{schulman2013finding,schulman2014motion} that exploits a heuristic function to remove potentially-interfering objects. This approach first plans a task and then tries to produce a motion plan that satisfies the computed set of discrete actions. If the induced motion planning problem is infeasible, the task planning is repeated taking into consideration the set of preconditions that identifies the infeasibility.,

Finally, Dantam et al.~\cite{dantam2016incremental} propose an incremental task and motion planner which combines discrete decisions about objects and actions with geometric decisions about collision free motion: they use an incremental constraint solver that adds motion constraints to the computed candidate task plan. The task plan is computed by using a Satisfiability Modulo Theories (SMT) approach, while the Open Motion Planning Library (OMPL) is used to find a feasible motion plan. At each failure, the algorithm iteratively increases the plan depth and motion planning timeouts such that it guarantees probabilistically completeness for fixed placements and grasps.

Instead of only focusing on manipulation domains, the proposed approach aims to integrate task and motion planning routines while performing generic tasks, that means robots should be able to use this system in order to handle both navigation and manipulation domains. Moreover, while most of the state-of-the-art approaches evaluate the objects relocation only when free-space motion planning is unfeasible, the proposed algorithm revalues a plan every time an action can save effort, not only when the ongoing trajectory becomes unfeasible.

\section{Problem Statement}
\label{sec:problem}

This Section defines the Deterministic Task (\ref{TP}) and Motion Planning (\ref{MP}) problems. These concepts will be the definition basis of the Conditional Task and Motion Planning authors will introduce in \ref{TMP}.

\subsection{Deterministic Task Planning}\label{TP}
Suppose the assignment of a task $T$ to a robot $R$. A task planner $\mathrm{TP}:(s_\mathrm{0}, s_\mathrm{G}, A) \to p^*$ aims to find an \textit{optimal} plan $p^* \in P$ solving $T$. $p^*$ moves $R$ from its start state $s_0 \in S$ to a goal state $s_G \in S$ by combining the set of actions $A$ that $R$ is able to perform according to its capabilities. 

The problem is deterministic if the actions domain is fully observable and every action $a \in A$ is fully defined as a sentence in the Planning Domain Definition Language (PDDL)~\cite{schulman2014motion} with a set $\mathrm{precon}(a)=\{\mathrm{precon}_\mathrm{0}(a), ..., \mathrm{precon}_\mathrm{N}(a)\}$ of preconditions 
and a set $\mathrm{effect}(a)=\{\mathrm{effect}_\mathrm{0}(a), ..., \mathrm{effect}_\mathrm{M}(a)\}$ of effects, 
described as conjunctive lists of literals in first-order logic.

$\mathrm{TP}$ computes a set of plans $P$, where $ p \in P$ is defined as
\[
 \quad p = \langle s_\mathrm{0}, a_\mathrm{0}, ..., s_{\mathrm{N}-1}, a_{\mathrm{N}-1}, s_\mathrm{N} \rangle, \quad s_\mathrm{N}=s_\mathrm{G}
\]
and $(s_i, a_i)\rightarrow s_{i+1}$ iff $\mathrm{precon}(a_i)$ is satisfied by $s_i$ and $\mathrm{effect}(a_i)$ brings to $s_{i+1}$. 

A plan $p^* \in P$ is \textit{optimal} if it has the lowest cost among all the computed plans:
\[
p^* = \mathrm{argmin}_{p \in P} \sum_{\langle s,a \rangle \in p} \mathrm{Cost}(\langle s,a \rangle)
\]
$\mathrm{Cost}(\langle s,a \rangle)$ is the cost of action $a$ being executed in state $s$.

\subsection{Deterministic Motion Planning}\label{MP}
 A motion planner $\mathrm{MP}:(s_\mathrm{0}, s_\mathrm{G}, A) \to t^*$ tries to find an \textit{optimal} path $t^* \in \tau$ that lets $R$ move from $s_0 \in S$ to $s_G \in S$ while avoiding collisions. The problem is deterministic if the working space is fully observable. In this case, MP can find a set of paths $\tau$, where $t \in \tau$ is a path in the free space:
 \[
 \tau : [0,1] \to C_\mathrm{free}, \quad \tau(0)=s_0, \quad \tau(1)=s_\mathrm{G}
\]
$t^*$ is \textit{optimal} if its trajectory is of minimum length:
\[
t^* = \mathrm{argmin}_{t \in \tau} \Big(\mathrm{Length}(t)\Big)
\]

\subsection{Conditional Task and Motion Planning}\label{TMP}
Suppose the existence of a Task and a Motion Planner (See \ref{TP} and \ref{MP}). Suppose that $T$ is assigned to $R$. $\mathrm{TMP}:(s_\mathrm{0}, s_\mathrm{G}, A) \to t^*$ finds the \textit{optimal} plan $p^* \in P$ performing $T$ and returns the \textit{optimal} trajectory $t^* \in \tau$ executing $p^*$. The solution is \textit{optimal} if $t^*$ is of minimum cost:

\[
t^* = \mathrm{argmin}_{p \in P} \Big(\sum_{0 \leq i \leq |p|} \mathrm{Cost}(t_i | a_i)\Big)
\]

where $\mathrm{Cost}(t_i | a_i)$ is the cost of the trajectory necessary to perform the i-th planned action. Without loss of generality, in this paper the solution is optimal if it minimizes the robot's effort: $\mathrm{Cost}(t_i | a_i) = \mathrm{Effort}(t_i | a_i)$ and $\mathrm{Effort}(t | p) = \sum_{0 \leq i \leq |p|} \mathrm{Effort}(t_i | a_i)$. The effort is defined as the execution time.

The problem is deterministic if the actions space is a priori fully defined and each action is executed infallibly. However, in the real world actions can generate unexpected effects and the robot can perceives changes at its surroundings. This means that the outcomes of environment and actuation actions should be processed in order to address uncertainties due to partial observability at the time of offline planning~\cite{nouman2016experimental}. The definition of $t^*$ is unchanged but the way used to find it is new: the plan $p^*$ should handle every known condition through the definition of a \textit{reachability graph} and an online recovery procedure should handle unexpected events by combining sensing and actuation actions and minimizing the global cost of the computed path.

\section{Algorithm}
\label{sec:algorithm}

\begin{figure*}[t!]
    \centering\includegraphics[scale=0.44]{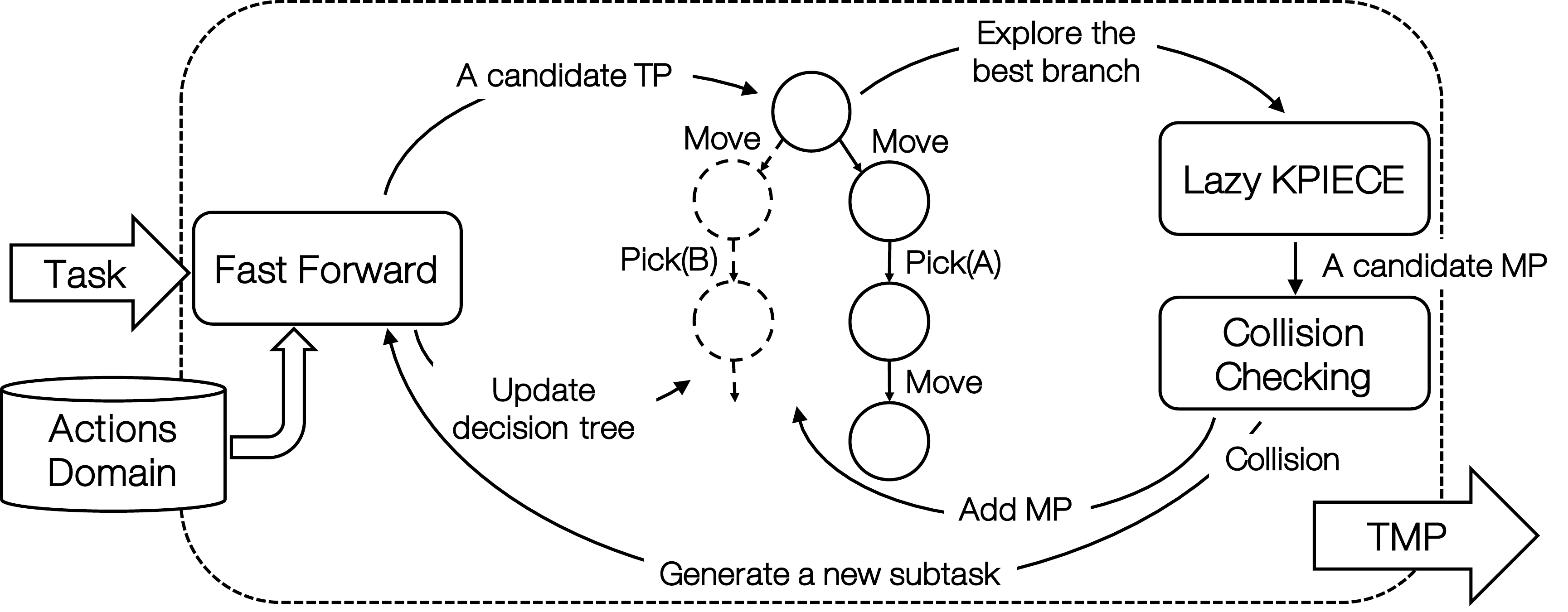}
    \caption{Algorithm Pipeline. The algorithm computes a candidate Task Plan (TP), explores the best branch, and computes the corresponding candidate Motion Plan (MP). It checks for collisions in the candidate Motion Plan. If a collision exists, it generates a new subtask to handle the collision object, it replans, and updates the \textit{reachability graph}. If there is no collision, it updates the \textit{reachability graph} with the MP. It repeats the pipeline until the goal state is reached.}
    \label{fig:schema}
\end{figure*}

A task $T$ is assigned to a robot $R$. $T$ asks $R$ to reach a goal state $s_G$. The complete set of actions $A$ that $R$ can perform and its initial state $s_0$ are known. Actions are expressed in PDDL. Starting from this information, Algorithm~\ref{alg:TMP} is applied.  

Let: - $P$ be the set of plans $\{p_0, ..., p_N\} \in P$ found by recursively applying the TP (FF) when an online replanning is necessary; -  $Obstacles$ be the set of objects encountered during the robot motion; - $G$ be the \textit{reachability graph} having as edges the actions of $P$ and as nodes the states of $P$. More in detail, every edge models the motion path used to execute that action. Initially, all these sets are empty. 

Algorithm~\ref{alg:TMP} starts by applying the TP (FF) and computing one sequence of possible actions letting $R$ accomplish $T$: $p = \langle s_\mathrm{0}, a_\mathrm{0}, ..., s_{\mathrm{i}}, a_{\mathrm{i}}, s_{\mathrm{i}+1}, ..., s_{\mathrm{G}} \rangle \in P$. The plan may not be optimal. Starting from $P$, Algorithm~\ref{alg:ktree} initiates $G$ with the set of actions to be performed and the list of states that are consequently reached.
 
Given the sequence of actions to be performed, \cite{castaman2016sampling} (a variation of the Lazy KPIECE) is used to compute the motion path $t^*$ connecting, when possible, every couple of states of $G$ through a motion trajectory not checked for collisions. The feasibility of the connection is checked in terms of actions' preconditions and effects. The aim is driving the robot towards the shortest path, in terms of Euclidean distance to the goal (see~\cite{castaman2016sampling}). If the environment is unknown, $t^*$ is computed until the last visible state. If $p^* \in G$ expects the execution of a navigation action, $t^*$ is computed in terms of the mobile base reference system. If it expects a manipulation action, $t^*$ combines the motion of the base with that of the manipulator robot. Moreover, \cite{ten2017grasp} is used to detect all possible 2-fingers grasps and a geometric variation of it should be used to detect 3-fingers grasps. 

Starting from $t^*$, the sequence of control inputs letting $R$ perform the trajectory is computed. This sequence lets deduce the effort $c^*$, in terms of costs, needed by the robot to perform motions. In the case in analysis, the algorithm takes into consideration the execution time (see ~\cite{castaman2016sampling}).

\begin{algorithm}[t!]
\KwIn{$s_0$: Start state; $s_G$: Goal state; $A$: Set of actions that $R$ can do}
\KwOut{$(t^*, c^*)$: Path of minimum cost letting $R$ execute the best plan}
$P$ = \{\}\tcp*[r]{set of plans}
$Obstacles$ = \{\}\tcp*[r]{list of encountered obstacles}
Initialize an empty reachibility graph $G$\;
$p \leftarrow \mathrm{TP}(s_0, s_\mathrm{G}, A)$\;
$P$.pushBack($p$)\;
$t^* \leftarrow \mathrm{LazyKPIECE}(s_0, s_\mathrm{G}) | p$\;
$c^* \leftarrow \mathrm{Effort}(t^* | p)$\;
$G^* \leftarrow \mathrm{RGraph}(P)$\;
$node^* \leftarrow G.root()$\;
    $cost$ = 0\;
    $t$ = \{\}\;
    Traverse($node$)\;
\textbf{return} $(t^*, c^*)$\;
\caption{TMP algorithm}
\label{alg:TMP}
\end{algorithm}

\begin{algorithm}[t!]
\KwIn{$P$: the set of plans}
\KwOut{$G$: the reachability graph of P}
\ForEach{state $s_i \in P$}{
    \If{$(s_i, a_i) \rightarrow s_{i+1}$}{
        $s_i$.children[].pushBack($s_{i+1}$)\;
        $s_{i+1}$.parent[].pushBack($s_i$)\;
    }
}
\caption{RGraph($P$)}
\label{alg:ktree}
\end{algorithm}

\begin{algorithm}[t!]
\KwIn{$v$: the node to be expanded}
\While{($v$.hasChild()) \&\& !(MaxAttempts reached)}{
    $t_\mathrm{Lazy} \leftarrow \mathrm{LazyKPIECE}(v, child)$\;
    \While{$t_\mathrm{Lazy}$ has new $collision \in C_\mathrm{movable}$}{
            $obj \leftarrow \mathrm{findCollisionObject}(collision)$\;
            \If{$(obj_\mathrm{label}, obj_\mathrm{pose}) \notin Obstacles$}{
                $Obstacles$.pushBack($obj$)\;
            }
            $A_{obj} \leftarrow \mathrm{findPossibleActions}(obj)$\;
            \ForEach{$a \in A$}{
                $S_\mathrm{effect} \leftarrow \mathrm{findEffectStates}(a)$\;
                \ForEach{$s \in S_\mathrm{effect}$}{
                    \If{$\exists p_\mathrm{before} \leftarrow \mathrm{TP}(v, s, A)$}{
                        \ForEach{$child$ of $v$}{
                            \If{$\exists p_\mathrm{after} \leftarrow \mathrm{TP}(s, child, A)$}{
                                $G \leftarrow \mathrm{updateRGraph}(G, p_\mathrm{before})$\;
                                $G \leftarrow \mathrm{updateRGraph}(G, p_\mathrm{after})$\;
                            }
                        }
                    }
                }
            }
    }
    $t \leftarrow t + \mathrm{KPIECE}(v, child)$\;
    $cost \leftarrow cost + \mathrm{Effort}(v, child)$\;
    \If{$cost < c^*$}{
        \eIf{$v == s_\mathrm{G}$}{
            $c^* \leftarrow cost$\;
            $t^* \leftarrow t$\;
            \textbf{return}\;
        }{
            Traverse($child$)
        }
    }
}
\caption{Traverse($v$): Expand $v$ to find the best $t^*$}
\label{alg:traverse}
\end{algorithm}
Being the robot at its current state $s_i$ (i.e., $s_0$ at the very beginning), the graph expansion starts (see Algorithm~\ref{alg:traverse}).
Given $t^* = t_\mathrm{Lazy}$, every node $s_i \in G$ and every edge $(s_i, s_{i+1}) \in G$ are checked for collisions. Authors remember that $G$ already contains the best plan performing the assigned task: the one which motion path has minimum Euclidean distance to the goal. Let the robot be at state $s_i$, for every new collision detected in the space of movable obstacles $C_\mathrm{movable}$, a state $s_\mathrm{j} \in S$ is sampled. Based again on the set of feasible actions $A$, their preconditions, and effects, $G$ is extended by adding the task plans from/to $s_\mathrm{j}$ (it is a new iteration of FF). 

The algorithm evaluates all the task plans $\langle s_\mathrm{i}, a_\mathrm{i}, ..., s_\mathrm{j}, a_\mathrm{j}, ..., s_{\mathrm{i}+N}, s_{\mathrm{G}} \rangle$ that connect the robot current state to the current plan passing through $s_j$. The related motion paths are evaluated too.
In detail, starting from the original path $\langle s_\mathrm{i},..., s_\mathrm{G} \rangle$, if the new path $\langle s_\mathrm{i}, ..., s_\mathrm{j} \rangle$ has a cost $c^*(s_\mathrm{i}, s_\mathrm{j})$
 less than $c^*(s_i, s_{G})$, the algorithm continues the exploration of this branch until reaching $s_\mathrm{G}$ or until exceeding $c^*(s_i, s_{G})$. In this case, continuing this road is no longer convenient and the exploration must focus on other branches. The exploration proceed until a better solution is found, all children have been visited, or the maximum number of attempts has been reached. This means that $p$ is revalued not only when unfeasible, but every time better solutions exist with respect to the selected one. For example, once an obstacle is found and a path avoiding it is computed, the algorithm evaluates both the action of avoiding the object and the one of manipulating it. It then selects the alternative of minimum effort, that in the case in analysis means the one requiring the minimum execution time. 

Figure~\ref{fig:schema} depicts the proposed algorithm pipeline, starting from the desired task assignment to the TAMP solution. The online replanning routine is also depicted; its details follow.

\subsection{On-Line Replanning}

\begin{figure*}[t!]
    \begin{subfigure}[t]{.32\linewidth}
        \centering\includegraphics[scale=0.55]{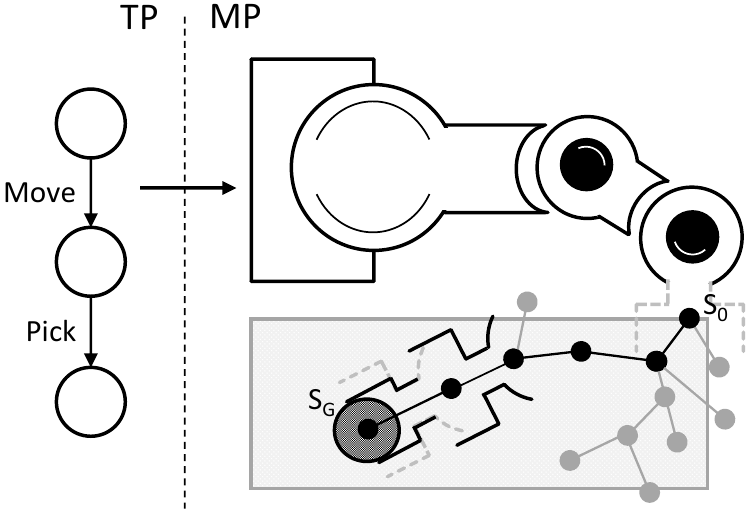}
        \caption{TP calculates a sequence of actions that achieves the assigned task. MP computes the corresponding sequence of motions.}
        \label{fig:replan_a}
    \end{subfigure}
    \hfill
    \begin{subfigure}[t]{.32\linewidth}
        \centering\includegraphics[scale=0.55]{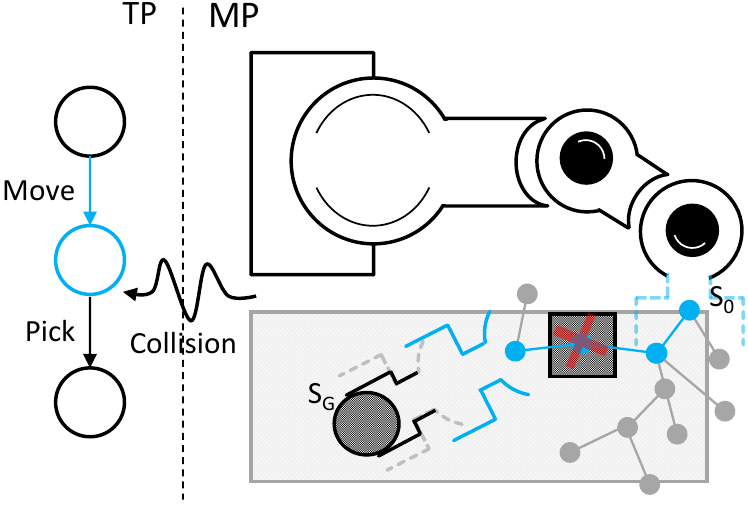}
        \caption{During the execution of the motion plan, the robot perceives a new obstacle.}
        \label{fig:replan_b}
    \end{subfigure}
    \hfill
    \begin{subfigure}[t]{.32\linewidth}
        \centering\includegraphics[scale=0.55]{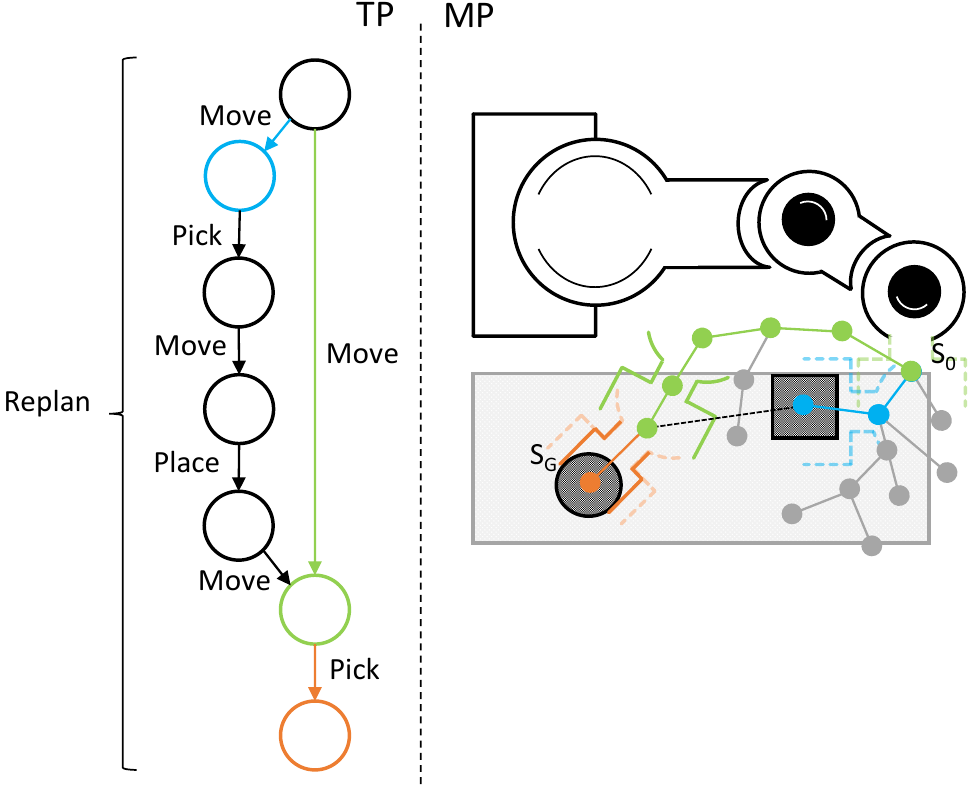}
        \caption{TP samples new states and tries to connect them to the existing plan. If a connection exists, a new sequence of actions is generated, and the less onerous branch is expanded.}
        \label{fig:replan_c}
    \end{subfigure}
    
    \caption{Online Replanning: a robot perceives a new obstacle while trying to pick up a known object. On the left of each figure, the plan of the Task Planner (TP) is depicted. On the right, the Motion Planner (MP)'s search space is visible.}
    \label{fig:replan}
\end{figure*}

Suppose the robot $R$ is executing a planned trajectory $t^*$, coming from an action $a \in p^*$, and in the meanwhile it continuously perceives its surrounding.
While acting in the real world, unpredictable faults could happen. E.g., the manipulated object may fall down from the gripper or an underestimated effort may be used to move the object itself. Moreover, a new obstacle $(\mathit{obj}_\mathrm{new_\mathrm{label}}, \mathit{obj}_\mathrm{new_\mathrm{pose}}) \notin \mathit{Obstacles}$ blocking the way could be perceived by sensors.
In all these situations, a task and path replanning is required. It should consider the new updated knowledge of the robot's world.

The online replaning procedure exploits the same method described above to add a new sub-task to the \textit{reachability graph}. Indeed, if the ongoing action fails, a new sub-task bypassing the faulty state is generated. The sub-task is computed by the task planner using the set of actions that the robot is able to perform and it is added to the \textit{reachability graph}. This sub-task is naturally connected to the next non-faulty action node, so that the approach does not need to perform a total replanning from the actual state $s_\mathrm{i}$ to the goal state $s_\mathrm{G}$.
In the end, the $\mathrm{Traverse}(s_\mathrm{i})$ function is invoked in order to try to find a path that minimizes the execution time. A number of attempts is chosen a priori: if no plan is found and totally executed within those attempts, the system outputs a failure.

Figure~\ref{fig:replan} presents a possible scenario where a robotic hand has to pick up an object. The Task and Motion Planner elaborates a plan to achieve the assigned task in which the robot moves near the object and then picks it up, as in Figure~\ref{fig:replan_a}.
During the execution of the \textit{Move} action a new object that prevents the robot to accomplish the movement is perceived by the sensors (see Figure~\ref{fig:replan_b}).
The online replanning generates a new sub-task corresponding to the obstacle replacement and adds it in the \textit{reachability graph}, as in Figure~\ref{fig:replan_c}.
The algorithm explores the graph and expands the less onerous branch that, in this example, corresponds to the elusion of the obstacle. As expected, the subsequent \textit{Pick} action is exactly the same for both suitable branches.

\subsection{Conditional Planning}

\begin{figure}[t!]
        \centering\includegraphics[scale=0.4]{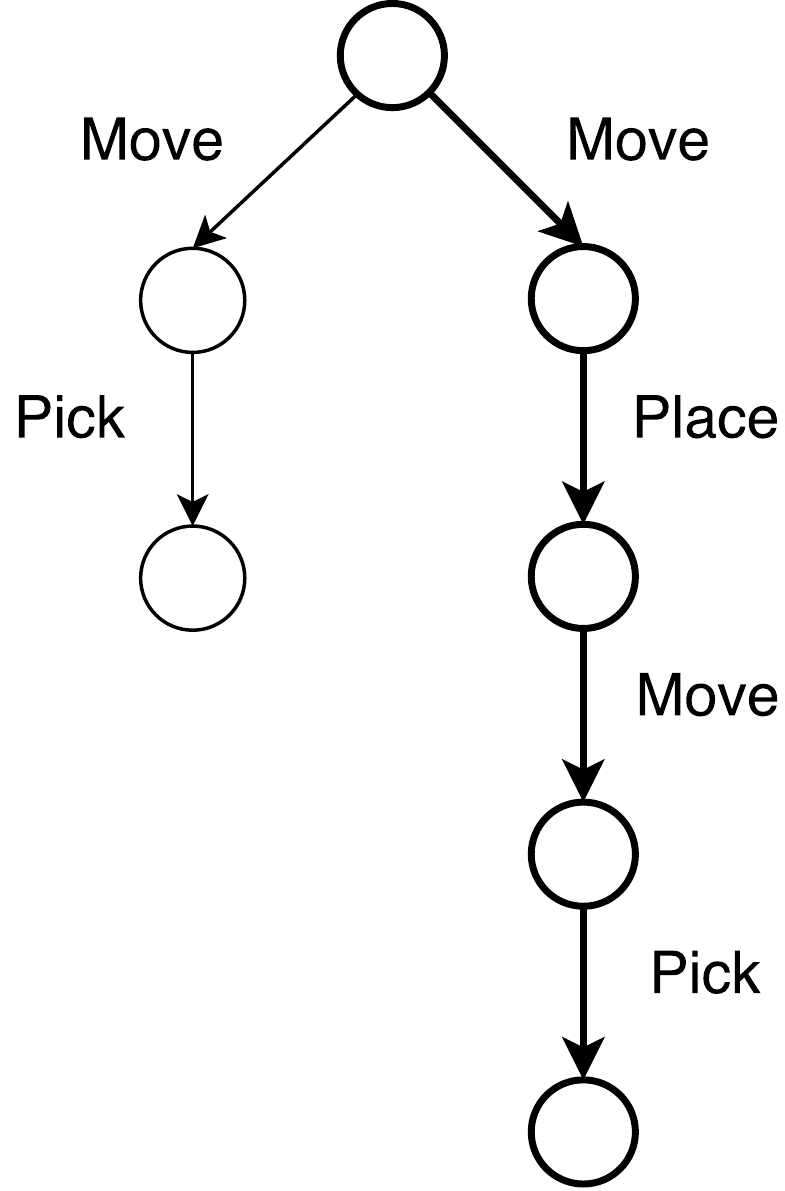}
        \caption{Conditional Planning: the robot has to pick an object. The figure depicts the $\mathit{Pick}$ \textit{reachability graph} generated in accordance with the satisfied preconditions. On the left, there is the sequence of actions to be performed when the gripper is empty. On the right, there is the sequence of actions to be performed when the gripper is not empty and the object it is holding has to be placed down before executing a new pick.}
        \label{fig:conditional}
\end{figure}

Suppose  $R$ is executing a trajectory $t^*$, coming from an action $a \in p^*$, and meanwhile it is perceiving its surrounding. Contemplated events may occur. E.g., the robot has to manipulate an object but its gripper is not empty or the object is occluded. From the literature, these events are treated by \textit{Condition Planning} routines. In the case in analysis, they are already handled by the planner through the construction of the \textit{reachibility graph}. No replanning procedure is involved, just the right sequence of actions is chosen (see Figure~\ref{fig:conditional}). This approach's capability justifies the choice of authors to call the proposal as a \textit{Conditional} approach.

\section{Experiments}
\label{sec:experiments}

\begin{figure}[t!]
    \centering
    \begin{subfigure}[t]{.49\linewidth}
        \centering\includegraphics[width=4cm]{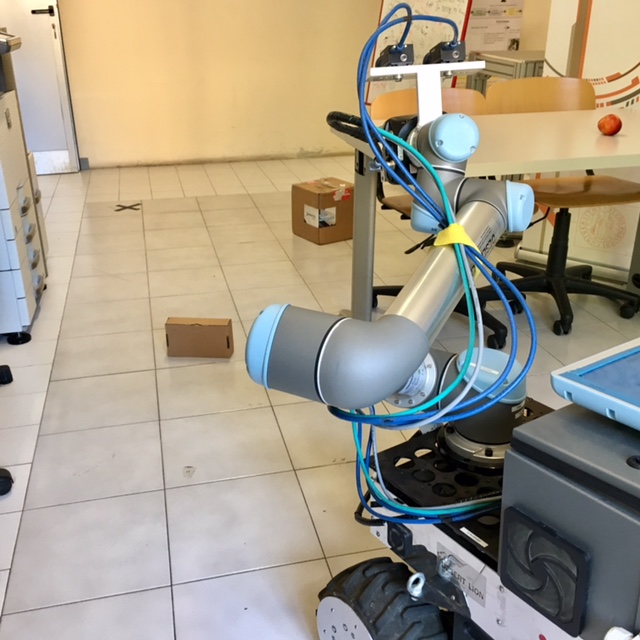}
        \caption{A mobile manipulator robot trying to solve a NAMO problem.}\label{fig:namo}
    \end{subfigure}
    \hfill
    \begin{subfigure}[t]{.49\linewidth}
        \centering\includegraphics[width=4cm]{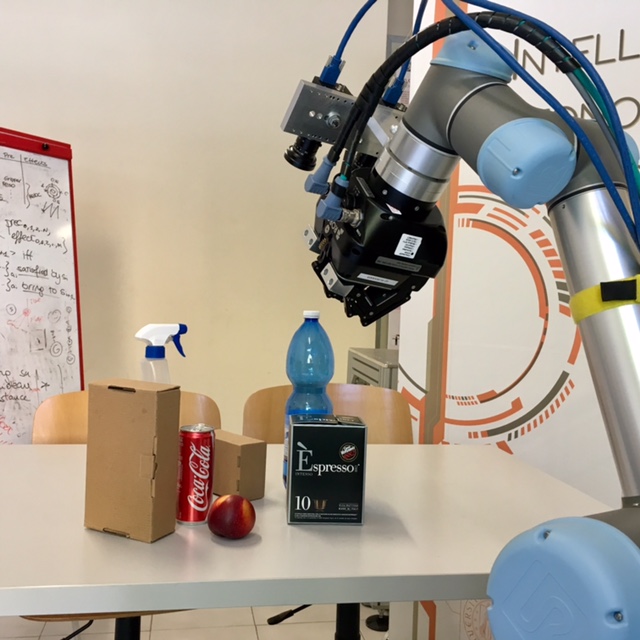}
        \caption{The same robot trying to pick up an occluded can of coke from a cluttered table.}\label{fig:manipulation}
    \end{subfigure}
 \caption{Use cases.}
 \label{fig:usecases}
\end{figure}

Figure~\ref{fig:usecases} depicts the use cases authors are studying at the time of the submission. Figure~\ref{fig:namo} shows a mobile manipulator robot trying to solve a NAMO problem. Figure~\ref{fig:manipulation} shows the same robot trying to pick up an occluded can of coke from a cluttered table. The robot can perform four different actions: 
\begin{itemize} 
\item $\mathit{Move}_\mathrm{base}(\mathit{pose}_\mathrm{start}, \mathit{pose}_\mathrm{goal}, \mathit{traj})$;
\item $\mathit{Move}_\mathrm{arm}(\mathit{pose}_\mathrm{start}, \mathit{pose}_\mathrm{goal}, \mathit{traj})$; 
\item $\mathit{Pick}(\mathit{obj}, \mathit{gripper}, \mathit{pose}_\mathrm{gripper}, \mathit{pose}_\mathrm{obj}, \mathit{conf}_\mathrm{joints}, \mathit{traj})$; 
\item $\mathit{Place}(\mathit{obj}, \mathit{gripper}, \mathit{pose}_\mathrm{gripper}, \mathit{pose}_\mathrm{obj}, \mathit{conf}_\mathrm{joints}, \mathit{traj},\\ \mathit{pose}_\mathrm{goal})$.
\end{itemize}
If $\mathit{pose}_\mathrm{goal}$ is not given as input, $\mathit{Move}$ randomly samples it on the free space and $\mathit{Place}$ does the same on a flat surface in the neighborhood of the manipulated object. 

Experiments aim to prove:
\begin{enumerate}
\item the adaptability of the algorithm when dealing with a perceived workspace;
\item the effectiveness of weighing paths based on the effort done.
\end{enumerate}
Authors consider the effort as the time spent and they aim to prove that the obtained solution is the fastest one.

\section{Conclusion} 
\label{sec:conclusion}
Authors presented a new algorithm able to solve a Task and Motion Planning problem through an effort-based approach. The effort is the time spent to accomplish the task and the algorithm finds the plan that can be executed in the shortest possible time. The non-determinism of the real world is faced by providing a Conditional Planning and a recovery routine that handles unexpected events and new scene detections. The proposed algorithm is complete and scalable.

Authors expose some use cases whose implementation is still in progress. They aim to prove the adaptability and effectiveness of the proposed approach.

\bibliographystyle{IEEEtran}
\bibliography{references}

\addtolength{\textheight}{-12cm}   


\end{document}